\documentclass{article}

\usepackage[preprint]{neurips_2024}

\usepackage[utf8]{inputenc} 
\usepackage[T1]{fontenc}    
\usepackage{hyperref}       
\usepackage{url}            
\usepackage{booktabs}       
\usepackage{amsfonts}       
\usepackage{nicefrac}       
\usepackage{microtype}      
\usepackage{xcolor}         

\usepackage{wrapfig}
\usepackage{soul}
\usepackage{caption}
\usepackage{graphicx}
\usepackage{amsmath}
\usepackage{amsthm}
\usepackage{algorithm}
\usepackage{graphicx} 
\usepackage{color}
\usepackage{dblfloatfix}
\usepackage{balance}
\usepackage{amssymb}

\usepackage{algorithm}
\usepackage{algorithmicx}
\usepackage{algpseudocode}

\usepackage{subcaption}
\usepackage{bm}
\usepackage{multirow}
\usepackage{enumitem}
\usepackage{algorithm}
\usepackage{algorithmicx}
\usepackage{algpseudocode}

\newcommand{\stitle}[1]{\vspace{0.5mm} \noindent {\bf #1}}
\renewcommand{\vec}[1]{\ensuremath{\mathbf{#1}}}

\newcommand{\bL}{\ensuremath{\mathcal{L}}}

\usepackage{enumitem}

\renewcommand{\vec}[1]{\ensuremath{\mathbf{#1}}}

\usepackage{multirow}

\newcommand{\model}{H$^2$GFM}{}

\title{H$\vec{^2}$GFM: Towards unifying Homogeneity and Heterogeneity on Text-Attributed Graphs}

\author{%
    Trung-Kien Nguyen \\
    University of Southern California \\
    Los Angeles, CA 90089, USA\\
    \texttt{kientngu@usc.edu} \\
    \And
    Heng Ping \\
    University of Southern California \\
    Los Angeles, CA 90089, USA\\
    \texttt{hping@usc.edu} \\
    \And
    Shixuan Li \\
    University of Southern California \\
    Los Angeles, CA 90089, USA\\
    \texttt{shixuan.li.ece@usc.edu} \\
    \And
    Peiyu Zhang\\
    University of Southern California \\
    Los Angeles, CA 90089, USA\\
    \texttt{pzhang65@usc.edu} \\
    \And
    Nikos Kanakaris\thanks{The work does not relate to the author's position at Amazon.}\\
    Amazon Web Services \\
    Seattle, WA 98109, USA\\
    \texttt{nikosk@amazon.com} \\
    \And
    Nicholas Kotov \\
    University of Michigan - Ann Arbor \\
    Ann Arbor, MI 48109, USA \\
    \texttt{kotov@umich.edu} \\
    \And
    Paul Bogdan \\
    University of Southern California \\
    Los Angeles, CA 90089, USA\\
    \texttt{pbogdan@usc.edu} \\
  }

\begin{document}

\maketitle
\begin{abstract}
The growing interest and applications of graph learning in diverse domains have propelled the development of a unified model generalizing well across different graphs and tasks, known as the Graph Foundation Model (GFM). Existing research has leveraged text-attributed graphs (TAGs) to tackle the heterogeneity in node features among graphs. However, it primarily focuses on homogeneous TAGs (HoTAGs), leaving heterogeneous TAGs (HeTAGs), where multiple types of nodes/edges reside, underexplored. To enhance the capabilities and applications of GFM, we introduce \model\, a novel framework designed to generalize across both HoTAGs and HeTAGs. Our model projects diverse meta-relations among graphs under a unified textual space and employs a context encoding to capture spatial and higher-order semantic relationships. To achieve robust node representations, we propose a novel context-adaptive graph transformer (CGT), effectively capturing information from both context neighbors and their relationships. Furthermore, we employ a mixture of CGT experts to capture the heterogeneity in structural patterns among graph types. Comprehensive experiments on a wide range of HoTAGs and HeTAGs as well as learning scenarios demonstrate the effectiveness of our model.
\end{abstract}

\section{Introduction}

Graph is a ubiquitous structure to model a wide range of real-world applications, such as complex particle systems~\cite{wang2020biomorphic, vecchio2021structural,wu2025layer}, recommendation~\cite{wu2020comprehensive, wu2022graph}, e-commerce~\cite{wu2020comprehensive, wu2022graph}, protein-protein, nanoparticle-protein or drug-target interactions in biomedicine~\cite{cha2022unifying}, or biological neural networks~\cite{yin2020network,kagan2022vitro}. Recently, Graph Neural Networks (GNNs) have become state-of-the-art method for fundamental graph tasks, yet
they usually are trained in a supervised manner, where each model is trained on a specific task for each individual dataset. With
the growing popularity of graph learning in diverse domains and applications, it is desirable to develop a Graph Foundation Model (GFM), a model trained from broad data that learns robust transferable graph representations to generalize on unseen graph data and tasks. GFMs offer several advantages \cite{chen2024text, liu2025graph}, including efficient training and adaptation to new, unseen data and enhanced generalization through scaling across domains.

An essential hindrance for training GFM is heterogeneity in graph features, where the node/edge features can have various dimensions across graphs. Even when dimensions are consistent, their meanings may still differ, leading to unreliable transferables. To tackle this issue, recent studies have investigated the usage of text-attributed graphs (TAG) \cite{ofa, chen2024text}, where node/edge features are expressed in natural language. These textual descriptions are then encoded by LLMs, producing unified representations under language model space. This approach supports semantics alignment across graphs and preserves the relevance for learned transferables. The growing availability of text features in graph data, combined with the high-quality representations from advanced LLMs, open up new avenues for text-space GFM research.

While existing text-space GFMs have achieved promising results \cite{ofa, chen2024llaga}, they mostly focus on homogeneous TAGs where graphs contain single node/edge type. However, the graphs in several real-world applications are constructed heterogeneously, where multiple types of nodes and edges reside \cite{lin2024heterophily, wang2022survey}. The multitude of node and edge types in heterogeneous graphs effectively captures complex and high-order semantics and connections in real-world scenarios. Nevertheless, it also adds challenges to modeling and generalizing across graph datasets, holding research underexplored. Furthermore, real-world systems can process multiple tasks involving both homogeneous (homo-) and heterogeneous (hetero-) graphs. For instance, an e-commerce platform may have various tasks, such as predicting products that are bought together as link prediction on homo-graph, or recommending products to users from their collaborative shopping patterns as link prediction on hetero-graphs. For drug design and drug repurposing scenarios, there could be collaboration between homo-graph of a protein and hetero-graph of interacting proteins along with their potential side effects. There could be potentials for graphs on similar or relevant domains to transfer knowledge and complement each other between those graph genres. Thus, we aim to develop text-space GFM to generalize well on both homo- and hetero-geneous graphs, or TAGs. While node features can be unified under the language feature space, two major issues still remain in designing GFM for heterogeneous TAGs:  

\textit{First, how can GFM capture the heterogeneity in node and  relation type across graph datasets?} Unlike homogeneous graphs having a single type of node and edge, heterogeneous graphs carry different numbers of type and semantics across datasets. For instance, an academic graph may include node types such as "Author," "Paper," and "Venue," while an e-commerce graph can have "User," "Item," and "Review" nodes. Classic GNNs treat all nodes and relations as a single type and thus are suboptimal for heterogeneous graphs. Heterogeneous graph neural networks (HGNNs) \cite{wang2019heterogeneous, hu2020heterogeneous} usually leverage type information of each specific dataset in their encoding mechanism, thus they can not be directly applied to new datasets with different types. Self-supervised approaches on heterogeneous graphs \cite{dmgi, heco} are also designed to pretrain and fine-tune on the same graph, lacking the flexibility to adapt to new graphs. Recently, HiGPT \cite{higpt} proposes heterogeneous graph tokenizer and tuning for LLM to generalize across heterogeneous graphs. However, the higher-order relationships are not explicitly modelled, while requiring heavy computing. To tackle this, we employ a novel relation encoding, a simple yet effective approach to capture both spatial and higher-order semantic relations. We first project textual meta-relations between nodes, which includes both two endnodes and relation types for local heterogeneity, under unified language space. Then a context graph is constructed for a target node to capture its neighborhood semantics and structural heterogeneity. The multi-hop relationships between target node and context nodes are computed  by the context encoding based on their elemental meta-relation representations. 

\textit{Second, how can GFM capture the heterogeneity in structural patterns among graph genres?} Homogeneous graphs often follow homophily, where neighboring nodes share similar labels or characteristics \cite{mcpherson2001birds}. This may not necessarily hold for heterogeneous graphs, where nodes can connect with others from different types with their own distinct properties. Closer neighbors may not be useful to identify the target node as a farther one from the same type. Even on same graph, sub-regions may posess different properties, such as local homophily or local heterophily \cite{zeng2023mixture}. In this work, we propose a context-adaptive graph transformer (CGT), producing robust, information-rich node representations from both node features and multi-hop relation representations from context graph. Furthermore, it is challenging for a single architecture to generalize effectively to multiple graphs and new, unseen ones. A natural design is to integrate the capabilities of a set of models tailored to diverse sub-domains patterns. Leveraging the concept of Mixture of Experts \cite{shazeer2017outrageously, gmoe}, we employ a group of our CGT experts, where each expert specializes in a set of context graph patterns. Unlike existing work \cite{xia2024anygraph, liu2024one} which aims to assign experts on the whole graph level, our MoE dynamically selects graph experts for specific target nodes, considering their local heterogeneity and structural patterns from the context graph for expert selection. 

\textbf{Contributions.} Our contributions in this work are summarized as follows: 1) We investigate a novel problem of developing GFM to handle both homogeneous and heterogeneous TAGs, promoting GFM capabilities and versatility for a wide-range of  applications. 2) We propose a novel framework named \model\, leveraging a context encoding to capture spatial and high-order semantic relationships between nodes, and a mixture of context-adaptive graph transformers to learn diverse structural patterns in the context subgraphs. 3) Extensive experiments on several benchmark datasets are conducted to demonstrate the effectiveness of our model over comparative baselines.



\section{Related Work}

\stitle{Graph neural networks.} Classic GNNs typically employ message-passing paradigm to obtain low-dimensional embedding for a target node by aggregating information from their local neighborhoods \cite{gcn, gat, xu2018powerful}. On heterogeneous graphs, Heterogenenous Graphs Neural Networks (HGNNs) \cite{hu2020heterogeneous, wang2019heterogeneous, lv2021we} exploit type-based information (node/edge/meta-path) to model rich semantic heterogeneity, usually as type-aware variants of attention mechanism. While those methods often achieve superior performance, their parametric designs rely on the type information for each specific graph, causing them to be inapplicable to new graphs.

\stitle{Graph self-supervised learning.} Self-supervised learning (SSL) on graphs has been extensively studied to pretrain robust representations under limited data labels. Using concepts from information theory, contrastive approaches form contrastive views to train similar instances to be closer while dissimilar instances are pushed away \cite{hou2022graphmae, velivckovic2018deep, you2020graph}. Contrastive methods on heterogeneous graphs \cite{dmgi, heco} exploit contrastive views from graph patches, schema networks, or meta-paths to capture high-order semantics. However, they are still designed to pretrain and fine-tune on same dataset, limiting their abilities to generalize to new data.

\stitle{Graph Foundation Model.} Along with the rise of LLM and foundation models, GFM aims to develop a pretrained model to adapt well on new, unseen data and a wide range of tasks. It offers several benefits: first, it is more cost-efficient to optimize a single model to solve for multiple tasks and datasets than to train one model per task per dataset like typical supervised methods. Second, GFMs can be efficiently adapted to new incoming datasets or tasks by partial fine-tuning, rather than training new ones. Furthermore, by scaling across datasets and tasks, they are expected to generalize relevant patterns, such as datasets in the same domain, to boost performance. Prominent approaches for GFM leveraging GNNs and/or LLMs, which can be categorized as: \textit{(1) Graph-based models:} modelling GNNs as predictors. OFA \cite{ofa}, PRODIGY \cite{huang2023prodigy} propose a prompting paradigm with Node-Of-Interest (NOI) subgraph to unify the training on a variety of graph tasks. GFT \cite{wang2025gft} leverages computation trees, a subgraph-like structure as graph tokens for transferable patterns. OMOG \cite{liu2024one} adapts a mixture of models by pretraining a model bank where each model is trained for a specific graph. Meanwhile, AnyGraph \cite{xia2024anygraph} projects hand-crafted node features and graph structure into unified representations using SVD and employs a mixture of graph experts for cross-domain heterogeneity. \textit{(2) LLM-based models.} In line of work leveraging LLM as graph predictor, Llaga \cite{chen2024llaga} transforms graph data into structure-preserving template and projects into token embedding space. Unigraph \cite{he2024unigraph} pretrains LM and GNN on TAG data, and leverages graph instruction tuning for zero-shot transfer. Meanwhile, TEA-GLM \cite{wang2025llms} employs a linear projector to transform graph embeddings from GNN into graph token embeddings without tuning LLM. HiGPT \cite{higpt} proposes in-context heterogeneous graph tokenizer and instruction tuning to enhance LLM capabilities on heterogeneous graphs. For a comprehensive review on graph foundation model research, please refer to the survey \cite{gfmsurvey}. TSFGM \cite{chen2024text} provides thorough benchmarks and evaluation for training GFM on text-space graphs. 

\stitle{Mixture of Experts (MoE).} MoE \cite{shazeer2017outrageously, chen2022towards} approach comprises $M$ sub-networks (experts) such as MLP, where each network aims to specialize in a specific sub-domain. Based on the input, a gating (routing) mechanism is utilized to dynamically assign proper experts for the process. In graph learning, MoE has been explored to enhance the capabilities of GNNs in supervised settings \cite{gmoe, linkmoe} or tackle cross-domain data \cite{xia2024anygraph, liu2024one}.
\section{Preliminaries}

In this section, we provide definitions or review the fundamental concepts for this work, as well as our problem formulation.

\stitle{Text-attributed Graph}. We formulate text-attributed graph (TAG) as $G = (V,E, S, T, R)$, where $V, E$ are the set of nodes and edges, $S$ is the set of text information incorporated with nodes, $T$ and $R$ are the set of node and edge types, respectively. If a graph has a single type of node and edge: $|T| = |R| = 1$, we define it as homogeneous TAG (HoTAG), otherwise it is denoted as heterogeneous TAG (HeTAG). We denote the text description for a node $u$ as $s_u$ and its type as $t_u$. Similarly, the text description for an edge $e$ is $s_e$ and its type is $r_e$. 

\stitle{Meta-relation and Meta-path.} In a heterogeneous graph, a meta-relation \cite{hu2020heterogeneous} for an edge $e=(u,v)$ is a tuple $\langle t_u, r_e, t_v \rangle$, capturing the heterogeneity of local neighborhoods by specifying the two endpoints and their relation types. Meanwhile, a metapath is defined as a sequence of node and edge types \cite{sun2011pathsim}: $P=T_{1} \xrightarrow[]{R_{1}} T_{2} \xrightarrow[]{R_{2}}...\xrightarrow[]{R_{l}}T_{l+1}$, where $T_{i}\in T$ and $R_{i} \in R$. From that, a meta-path can be regarded as a composition of meta-relations, capturing diverse, high-order semantics and relations between distant nodes. 

\stitle{Problem formulation}. Given a set of graph tasks $T$ and a set of pretrain graph datasets $D^{tr}$, we denote a pretrain job as $J^{tr}_i = \{D^{tr}_j, T_k\}$ if the task $T_k$ is available for the graph $D^{tr}_i$. Similarly, a downstream job is defined as $J^{t}_m = \{D^{t}_n, T_k\}$ for the tasks $T_k$ and the graph $D^{t}_m$ from downstream graph datasets $D^{t}$. We train a GFM model $\theta$ on the total of available pretrain jobs $J^{tr}$. The trained model can then be applied or fine-tuned for the set of downstream jobs $J^{t}$. Here, a graph $D^{tr}_i$, $D^{t}_n$ could be HoTAG or HeTAG as defined. In this work, we focus on two prevalent graph tasks: (\textit{i}) node classification (NC) and  (\textit{ii}) link prediction (LP). While the NC task aims to predict the label for a target node, the LP task aims to predict whether a link exists between two target nodes.

To thoroughly evaluate GFM, we investigate two scenarios: \textit{co-training} and \textit{cross-training}. In \textit{co-training}, where $D^{tr} = D^t$, the trained model is applied to evaluate the testing split of the pretrain set. This scenario evaluates the model's generalizability among various datasets. In \textit{cross-training}, where $D^{tr} \neq D^t$, the trained model is applied to novel datasets unseen in the training stage. Specifically, we evaluate the GFM under two scenarios: \textit{zero-shot} and \textit{finetuning}. In \textit{zero-shot}, the model is directly applied without any adaptation. In  \textit{finetuning}, the model is adapted with available samples. 

\begin{figure*}[t]
    \centering
    \includegraphics[width=1.0\linewidth]{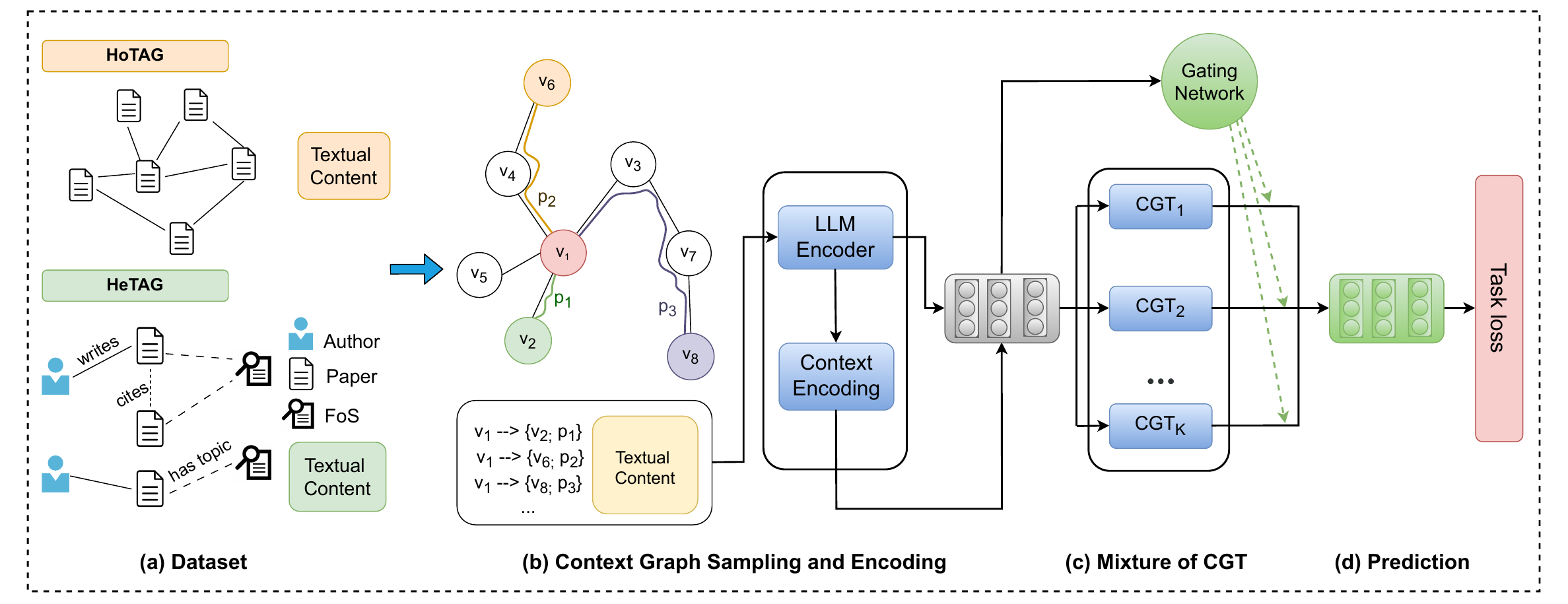}
    \caption{Overview framework of \model}
    \label{fig:framework}
\end{figure*}

\section{Proposed Method: \model}
In this section, we introduce the proposed method, \model, with its overall framework illustrated in Fig.\ref{fig:framework}. Given HoTAG/HeTAG datasets, a context graph is sampled for each target node, consisting of multi-hop, multi-relation neighbor nodes. We then employ an LLM and context encoding to obtain unified embeddings for node features and high-order relationships. Then, a mixture of context-adaptive graph transformers is applied to produce robust node embeddings for downstream tasks. We present each module as the following. 

\subsection{Context Graph Representation}
To unify the heterogeneity in node features among graphs, we leverage an LLM to encode textual features from TAG datasets into a unified language embedding space. Following existing works \cite{ofa, chen2024text}, we adapt Sentence-BERT \cite{reimers2019sentence} due to their performance and efficient computation. The embeddings $\vec{h}_u \in R^{d}$ for the node $u$ can be obtained as  
\begin{equation}
    \vec{h}_u =\textit{Sentence-BERT}({s}_u)    
\end{equation}
We also construct text description for each relation. For a given edge $e=(u,v)$, we generate its description using the template $s_e =$ "$t_u \, || \, r_e \, || \,t_v || s_e$". For instance, "A paper is written by an author." Additional available edge textual content can be concatenated to further enrich the semantics. We refer to this as textual meta-relation due to its alignment with meta-relation concept, efficiently capturing both relation and associated nodes' types for local heterogeneity. The resulting embeddings $\vec{r}_e \in R^{d}$ for edge $e$ can be generated from the text by LLM:
\begin{equation}
    \vec{r}_{e} = \textit{Sentence-BERT}(s_e)    
\end{equation}
\paragraph{Context Graph Sampling} To effectively capture heterogeneous and high-order information surrounding a target node $u$, we sample a set of meta-paths and retrieve their endnodes as context nodes -  multi-hop, multi-relation neighbors for $u$. The target node and its context neighbors, denoted as $N_u$, thus, form a \textit{context graph}, where edges are corresponding meta-paths. Selection and sampling of meta-paths for different graphs require manual design and domain-specific knowledge. To facilitate and generalize this process, we adopt random walk-based sampling with variable lengths $L \leq L_{max}$ as in \cite{nguyen2023link}. Specifically, we generate $N$ random walks $\{p_1,\ldots,p_{N}\}$ of maximum length $L_{\text{max}}$ for each target node, starting from itself. These paths are subsequently truncated to a shorter length $L \leq L_{\text{max}}$ to produce variable-length paths. Given that each path incorporates type information of all node/edge elements, it is equivalent to an instance of meta-path. 
This strategy enables the model to capture high-order semantic dependencies across multi-relation types and diverse structural ranges. Context graph can be regarded as a customized subgraph, inheriting several of its advantages. Notably, its sampling approach enables mini-batch training on large-scale graphs. It  also supports inductive setting where graph structure may evolve over time, making it well-suited for GFM applications.



Then it comes to the question of how to encode higher-order semantics from meta-paths? It appears impractical to enumerate all diverse types and lengths to feed into LLM for embeddings. Thus, we propose to encode them based on unified embeddings of their unit meta-relations. Given a meta-path consisting of subsequent meta-relations $p=\{e_1, e_2, .., e_L\}$, our context encoding incorporates both spatial relation and higher-order semantic relations into its embedding $\vec{h_p} \in R^d$ as follows:
\begin{equation}
    \vec{h}_{p} = \sum^L_{m=1} {\frac{1}{m}\vec{r}_{e_m}},
\end{equation}
where $\vec{r}_{e_m}$ denotes embeddings of the $m$-hops meta-relation from $u$. We discuss the expressiveness of our context encoding. On HoTAG with a single meta-relation, it distinguishes connections from context neighbors at various hops, unlike conventional pooling methods such as $\textsc{Mean}(.)$ or $\textsc{Max}(.)$. It also aligns with the homophily principle commonly observed in graphs, where closer relations tend to be more important and assigned with higher weights rather than uniform $\textsc{Sum}(.)$. On HeTAG, this encoding effectively captures diverse high-order relationships by aggregation of compositional relations. For instance, given two meta-paths P-A-P and P-A-P-A-P, context encoding is able to capture the differences, while $\textsc{Mean}(.)$ or $\textsc{Max}(.)$ would fail.

\subsection{Context-Adaptive Graph Transformers}


While a wide range of GNN architectures can be utilized to derive node representations for learning tasks, we specifically design a Context-adaptive Graph Transformer (CGT) to effectively capture the rich context and high-order relationships from the context graph. This design enables the model to leverage the high-order semantics in two critical aspects: (1) the interactions between the target node and context neighbors as attention scores, and (2) the transformation of context neighbors' information for aggregation. Given a target node $u$ and a context node $v \in N_u$ connected by meta-path $p$, the  $l$-th CGT layer takes node embeddings and meta-path embeddings into account to compute attention scores:
\begin{align}
\vec{h}^l_{att} &= [\vec{W}^l_q \vec{h}^{l-1}_u \,\Vert\, \vec{W}^l_k \vec{h}^{l-1}_v \,\Vert\, \vec{W}^l_p \, \vec{h}^{l-1}_p] \\
\alpha^l_{(u,v)} &= Softmax\big(\sigma(\vec{W}^l_a \vec{h}^l_{att}) \big),
\end{align}
where $\vec{W}^l_q, \vec{W}^l_k, \vec{W}^l_r\in\mathbb{R}^{d_h^{l}\times d_h^{l-1}}$, $\vec{W}^l_a\in\mathbb{R}^{1\times 3d_h^{l}}$ are the learnable parameters and $\sigma$ is the $\textsc{LeakyReLU()}$ activation. To adjust context neighbors' message regarding specific high-order relationship from their connected meta-path, we employ the transformation as linear feature-wise modulation (FiLM) \cite{film}:
\begin{align}
    {\vec{m}}^{l}_{v} = (\gamma^{l}_{p}+\vec{1})\odot (\vec{W}^l_v \vec{h}^{l-1}_v) + \eta^{l}_{p},\label{eq.personalization}
\end{align}
where $\vec{W}^l_v \in \mathbb{R}^{d_h^{l}\times d_h^{l-1}}$ is learnable parameters, $\vec{1}$ centers the scaling around one, and $\odot$ denotes element-wise multiplication. For the scaling and shifting vectors $\gamma^{l}_{p}, \eta^{l}_{p} \in R^{d_h^{l}}$,
are  conditioned on the meta-path embeddings. We implement the function as multilayer perceptron (MLP):
\begin{align}
    \gamma^{l}_{p}=\sigma(\vec{W}^{l}_{\gamma}\vec{h}^{l}_{p}+\vec{b}^{l}_{\gamma}), \quad \eta^{l}_{p}=\sigma(\vec{W}^{l}_{\beta}\vec{h}^{l}_{p}+\vec{b}^{l}_{\beta}),\label{eq.film}
\end{align}
where $\vec{W}^l_v, \vec{W}^l_u \in\mathbb{R}^{d_h^{l}\times d_h^{l-1}}$ are learnable parameters. Finally, embedding for target node $u$ in layer $l$-th is achieved by aggregating context neighbors' messages with attention weights: 
\begin{equation}
    \vec{h}^l_u = \sigma\Big[ \Big(\sum^{N_u}_{v}\alpha^l_{(u,v)}\vec{m}^{l}_{v}\Big) + \vec{W}^l_u\vec{h}^{l-1}_u \Big],
\end{equation}
where $\vec{W}^l_v, \vec{W}^l_u \in\mathbb{R}^{d_h^{l}\times d_h^{l-1}}$ are learnable parameters.



\paragraph{Mixture of CGT Experts} To tackle the heterogeneity in semantics and structure among graph types, we employ a mixture of our CGT experts, where each expert aims to excel for a set of local context graph patterns of the target node $u$. The final representation for $u$ is achieved by incorporating from top $k$ of total $n$ graph experts:
\begin{equation}
    \vec{h}^l_u = \sum^{k}_{i=1} G_i\big(\vec{h}^{l-1}_u\big) \cdot CGT_i\big(\vec{h}^{l-1}_u\big)
\end{equation}
We design the sparsely $TopK$ gating mechanism following \cite{shazeer2017outrageously}:
\begin{align}
     G_i(\vec{u}) &= Softmax(TopK(H(\vec{u})), k)) \\
    H(\vec{u})_i &= (\vec{W}_g \vec{u})_i +  \epsilon \cdot Softplus((\vec{W}_{\epsilon} \vec{u})_i) \\ 
    TopK(x, k)_i &=
    \begin{cases} 
    x_i & \text{if } x_i \text{ is in top } k \text{ elements of } x. \\
    -\infty & \text{otherwise}.
    \end{cases}
\end{align}
 where $\vec{W}_g, \vec{W}_{\epsilon} \in \mathbb{R}^{n \times d_h^{l-1}}$ are learnable parameters, noise $\epsilon \sim \mathcal{N}(\vec{0}, \vec{1})$ and $\vec{u}$ is short for $\vec{h}^{l-1}_u$. The learnable Gaussian noise is added before applying softmax facilitates the load-balancing, which is controlled by the weight $\vec{W}_{\epsilon}$. Then, the top \( k \) values are kept, and the rest are set to \( -\infty \), resulting in 0 value for the corresponding gate.









\subsection{Training Objective}

While prompt-graph methods \cite{ofa, wang2025gft} can handle different discrete classification tasks under a unified training scheme, they are impractical for regression tasks that predict continuous values. Alternatively, we adopt a plug-and-play strategy, where a task-specific predictor $\theta_i$ is designed for each downstream task. Predictors are designed to map robust learned embeddings into task prediction space, typically using simple designs such as an MLP layer. In this work, we leverage two fundamental graph tasks-node classification (NC) and link prediction (LP)--as the basis for (co-)training. For newly coming tasks in the later phase, we can flexibly select an appropriate predictor or design a new one with little fine-tuning. Under mini-batch training, the model is trained to optimize specific task $L_{mb-task}$ for each mini-batch samples:
\begin{equation}
    L_{mb-task} = \begin{cases} 
    -\sum y \cdot log\big[\theta_{NC}(\vec{h}_u, \vec{h}_y) \big] & \text{if task is NC} \\
    -\sum y \cdot log\big[\theta_{LP}(\vec{h}_u, \vec{h}_v) \big] & \text{if task is LP}.
    \end{cases}
    \label{eq:loss_task}
\end{equation}
where $y$ is the ground-truth and $\vec{h}_y$ denotes its textual embeddings. We formulate node classification as multi-class classification task, where prediction is produced from node and ground-truth embeddings. This enables zero-shot prediction on new, unseen datasets. Link prediction is formulated as binary classification, where the predictor estimates the probability if two nodes $(u, v)$ are connected. We further present the training algorithm for \model\ as well as complexity analysis in Appendix A.
\section{Experiments}



In this section, we perform extensive experiments of \model\ on several public benchmark datasets and comparative baselines. 
They evaluate model effectiveness in co-training on HoTAGs and HeTAGs, as well as its transferability to new datasets in zero-shot and fine-tuning settings. We then investigate the impact of each module in \model\  through ablation and parameter sensitivity analyses. 

\subsection{Experimental Setup}
\paragraph{Datasets} 
We utilize eight HoTAG datasets from \cite{chen2024text} and four HeHTAG datasets from \cite{liu2024multi}, which come from diverse domains such as academic network, e-commerce, etc. The statistics of our datasets and  scenario splits are summarized in Table. \ref{tab:dataset}, and further details are provided in Appendix B.
\begin{table}
\centering
\caption{Statistics of datasets.}
\label{tab:dataset}
\begin{tabular}{llrrll}
\toprule
\textbf{Setting} & \textbf{Dataset} & \textbf{\# Nodes} & \textbf{\# Edges} & 
\textbf{Type} & \textbf{Domain} \\ 
\midrule
\multirow{6}{*}{\textbf{Co-train}} 
& Arxiv & 169,343 & 2,315,598 & HoTAG &  CS Citation \\
& DBLP & 14,376 & 431,326 & HoTAG & CS Citation \\ 
& Products & 316,513 & 19,337,722 & HoTAG & E-commerce \\
& History & 41,551 & 503,180 & HoTAG & E-commerce \\
& He\_ArXiv & 231,111 & 2,075,692 & HeTAG & Academic \\ 
& He\_CroVal & 44,386 & 164,981 & HeTAG & CQA \\ 
\midrule
\multirow{6}{*}{\textbf{Cross-train}} 
& CiteSeer & 3,186 & 8,450 & HoTAG & CS Citation \\
& Pubmed & 19,717 & 88,648 & HoTAG & Bio Citation \\
& WikiCS & 11,701 & 431,726 & HoTAG & Knowledge \\
& Child & 76,875 & 2,325,044 & HoTAG & E-commerce \\
& He\_TMDB & 24,412 & 104,858 & HeTAG & Movie \\ 
& He\_DBLP & 1,989,010 & 28,390,033 & HeTAG & Academic \\ 
\bottomrule
\end{tabular}
\end{table}


\begin{table*}
    \centering
    \small
    \caption{Evaluation of node classification (NC) and link prediction (LP) in co-training setting. Best is bolded and runner-up is underlined. OOM denotes out-of-memory.}
    \label{tab.co-train}
    \begin{tabular}{l|cccccc}
        \toprule
        Tasks & Arxiv & DBLP & Products & History & He\_Croval & He\_Arxiv \\
        \midrule
        \midrule
        \multicolumn{7}{c}{\textbf{NC (ACC $\uparrow$)}} \\
        \midrule
        GraphMAE & 64.76 ± 0.40 & 62.83 ± 0.91 & 80.07 ± 0.28 & 80.42 ± 0.35 & 56.29 ± 0.32 & 74.69 ± 0.06 \\
        OFA & \underline{70.31} ± 0.32 & 65.63 ± 2.54 & \underline{85.42} ± 0.14 & \underline{83.10} ± 0.31 & 77.29 ± 0.95 & \underline{77.03} ± 0.83 \\
        LlaGA & 62.18 ± 0.32 & 28.61 ± 0.46 & 67.81 ± 0.77 & 47.64 ± 0.43 & 36.46 ± 0.61 & 42.82 ± 0.33 \\
        AnyGraph & 2.37 ± 1.25 & 26.15 ± 1.93 & 1.75 ± 0.98 & 1.50 ± 0.89 & 16.73 ± 2.12 & 1.33 ± 0.38 \\
        GFT & 68.52 ± 0.25 & \underline{72.23} ± 0.12 & OOM & 83.26 ± 0.15 & \underline{81.31} ± 2.57 & \underline{80.57} ± 0.04 \\
        \textbf{\model} & \textbf{77.10} ± 1.24 & \textbf{76.92} ± 1.08 & \textbf{86.77} ± 1.01 & \textbf{85.20} ± 0.97 & \textbf{84.20} ± 1.10 & \textbf{83.07} ± 1.03 \\
        \midrule
        Improve (\%) & 10.72\% & 5.86\% & 1.65\% & 2.89\% & 2.32\% & 2.58\% \\
        \midrule
        \midrule
        \multicolumn{7}{c}{\textbf{LP (AUC $\uparrow$)}} \\
        \midrule
        GraphMAE & 59.54 ± 1.11 & 78.95 ± 0.04 & 52.51 ± 2.48 & 59.19 ± 0.01 & 80.32 ± 0.41 & 63.09 ± 0.01 \\
        OFA & \underline{74.79} ± 0.19 & \underline{84.14} ± 1.59 & \underline{60.96} ± 0.19 & \underline{72.98} ± 0.19 & \textbf{98.28} ± 0.95 & 84.54 ± 1.03 \\
        AnyGraph & 54.65 ± 0.09 & 56.41 ± 0.12 & 59.29 ± 0.05 & 58.51 ± 0.03 & 44.81 ± 0.15 & 64.65 ± 0.11 \\
        GFT & 69.70 ± 0.17  & 77.07 ± 0.63  & 54.16 ± 0.27  & 61.38 ± 0.32 & 96.93 ± 0.10 & \underline{88.94} ± 0.07 \\
        \textbf{\model} & \textbf{75.10} ± 0.97 & \textbf{94.95} ± 1.05 & \textbf{69.77} ± 0.99 & \textbf{79.19} ± 1.02 & \underline{97.93} ± 0.96 & \textbf{91.50} ± 0.09 \\
        \midrule
        Improve (\%) & 0.41\% & 10.77\% & 14.15\% & 8.21\% & -0.36\% & 2.88\% \\
        \bottomrule
    \end{tabular}
\end{table*}

\paragraph{Baselines} We employ comprehensive baselines:  (1) \emph{Graph self-supervised learning}: GraphMAE \cite{hou2022graphmae}, (2) Graph-based foundation: OFA \cite{ofa}, Anygraph \cite{xia2024anygraph}, GFT \cite{wang2025gft} and (3) LLM-based foundation: LlaGa \cite{chen2024llaga}. We provide model details and parameters settings in Appendix C.  

\paragraph{Evaluation setting} 
In each dataset, we conduct node classification and link prediction tasks. For node classification, we use the benchmark train/test split from \cite{chen2024text,liu2024multi} as in Appendix B, and use accuracy (ACC) for evaluation metrics. The link prediction is conducted as binary classification, where we evaluate by AUC-ROC (AUC) metric. For each graph, we split the total number of edges by a ratio of 80\%/10\%/10\% for train/validate/test, and models are only trained on the subgraph constructed from train edges. For large datasets, we further sample a subset of 100k edge pairs from the train split, 10k pairs from each validate and test split and utilize them for the train/validate/test. We sample 2-hop neighbors as negative nodes for source nodes to maintain the task difficulty \cite{lv2021we}. The ratio of positive and negative edges is set as 1:1. We report results averaged over five runs.

\paragraph{Hyperparameter tuning} 
In our model \model, we sample $N=50$ meta-paths for 50 context neighbors and set the maximum length $L_{max}$ as 4. For LLM, we use Sentence-BERT \cite{reimers2019sentence} to generate textual embeddings of dimension 384. For the mixture of graph experts, we employ a total of 8 CGT experts, with 4 active experts for each prediction. Our CGT includes 1 layer with hidden dimension of 768 and output dimension of 384. The model is trained by the Adam optimizer \cite{diederik2014adam} with a learning rate of 0.001 and dropout ratio of 0.15. We provide further parameter details in Appendix D.

\vspace{-1mm}
\subsection{Evaluation on various scenarios}


\begin{table*}
    \centering
    \small
    \caption{Evaluation of zero-shot node classification (NC) and link prediction (PD) in cross-training setting. Best is bolded and runner-up is underlined.}
    \label{tab.zero-shot}
    \begin{tabular}{l|cccccc}
        \toprule
        Methods & Citeseer & Pubmed & Wikics & Child & He\_TMDB & He\_DBLP \\
        \midrule
        \midrule
        \multicolumn{7}{c}{\textbf{NC (ACC $\uparrow$)}} \\
        \midrule
        OFA & \underline{43.41} ± 0.95 & 24.98 ± 0.89 & \underline{10.31} ± 0.71 & 3.73 ± 0.62 & \underline{21.44} ± 0.78 & 11.78 ± 0.65 \\
        LlaGA & 11.91 ± 0.85 & 17.58 ± 0.74 & 13.72 ± 0.83 & \underline{4.40} ± 0.03 & 20.16 ± 0.79 & \underline{13.75} ± 0.68 \\
        AnyGraph & 18.55 ± 0.76 & \underline{35.19} ± 2.71 & 8.67 ± 0.82 & 1.60 ± 0.20 & 21.50 ± 1.28 & 6.27 ± 0.02 \\
        \textbf{\model} & \textbf{46.37} ± 0.98 & \textbf{38.46} ± 0.91 & \textbf{49.47} ± 1.03 & \textbf{6.58} ± 0.68 & \textbf{29.01} ± 0.85 & \textbf{18.72} ± 0.74 \\
        \midrule
       
        Improve (\%) & 6.81\% & 9.29\% & 379.83\% & 49.54\% & 35.31\% & 36.15\% \\
        \midrule        
        \midrule
        \multicolumn{7}{c}{\textbf{LP (AUC $\uparrow$)}} \\
        \midrule
        OFA & \underline{74.31} ± 0.93 & 36.95 ± 0.82 & 50.61 ± 0.87 & 59.45 ± 0.92 & \underline{74.94} ± 0.95 & \underline{76.86} ± 0.89 \\
        AnyGraph & 50.23 ± 0.12 & \underline{37.03} ± 0.04 & 59.76 ± 0.03 & 59.63 ± 0.09 & 36.24 ± 0.07 & 63.72 ± 0.11 \\
        \textbf{\model} & \textbf{81.15} ± 0.98 & \textbf{63.31} ± 0.91 & \textbf{68.99} ± 0.95 & \textbf{67.78} ± 0.96 & \textbf{97.13} ± 0.89 & \textbf{82.55} ± 0.93 \\
         \midrule
        Improve (\%) & 9.20\% & 70.97\% & 15.45\% & 13.67\% & 29.61\% & 7.40\% \\
        \bottomrule
    \end{tabular}
\end{table*}


\begin{table*}
    \centering
    \footnotesize
    \caption{Evaluation of fine-tuning node classification (NC) in cross-training setting. Best is bolded and runner-up is underlined. OOM denotes out-of-memory.}
    \label{tab.fine-tune}
    \begin{tabular}{l|cccccc}
        \toprule
        Methods & Citeseer & Pubmed & Wikics & Child & He\_TMDB & He\_DBLP \\
        \midrule
        \midrule
        GraphMAE & 69.88 ± 0.34 & 69.99 ± 0.98 & 65.50 ± 0.23 & 45.78 ± 0.85 & 55.20 ± 0.61 & 69.32 ± 0.27 \\
        OFA & 66.35 ± 0.32 & 67.88 ± 0.48 & 68.51 ± 1.02 & \underline{54.60} ± 0.95 & 75.27 ± 1.01 & \underline{73.26} ± 0.13 \\
        LlaGA & 13.68 ± 0.28 & 35.34 ± 0.47 & 25.26 ± 0.65 & 10.13 ± 0.36 & 22.08 ± 0.29 & 18.39 ± 0.09 \\
        AnyGraph & 16.67 ± 0.79 & 32.98 ± 1.91 & 11.51 ± 0.85 & 2.01 ± 0.68 & 23.77 ± 0.87 & 19.75 ± 0.90 \\
        GFT & \underline{72.88} ± 0.48  & \underline{70.19} ± 0.98 & \underline{78.25} ± 1.04 & 53.52 ± 0.97 & \underline{79.47} ± 1.08 & OOM \\
        \textbf{\model} & \textbf{73.68} ± 0.35 & \textbf{75.26} ± 1.01 & \textbf{79.19} ± 1.05 & \textbf{56.89} ± 0.98 & \textbf{79.68} ± 1.06 & \textbf{73.68} ± 1.02 \\
        \midrule
        Improve (\%) & 1.10\% & 7.22\% & 1.20\% & 4.19\% & 0.26\% & 0.57\%  \\
        \bottomrule
    \end{tabular}
\end{table*}

\stitle{Cross-training.} We report the results for the co-training setting in Table. \ref{tab.co-train}. Overall, our proposed model significantly outperforms other comparative baselines on both node classification and link prediction tasks. The results imply that \model\ can capture diverse, high-order semantics and structural patterns to effectively generalize among various HoTAGs and HeTAGs. We make further observations. AnyGraph does not perform well on node classification, potentially due to their training objective, which is primarily designed for link prediction. OFA is more robust yet remains suboptimal, perhaps because their single graph encoder lacks the flexibility to capture heterogeneous characteristics of various datasets. GFT can achieve consistent performance, yet they still require fine-tuning for each dataset and do not scale to large datasets. 

\stitle{Cross-training.} We leverage the pretrained model from co-training phase to evaluate its adaptability for unseen datasets in cross-train set. We report results for the zero-shot setting in Table.\ref{tab.zero-shot}, where no data samples are provided for fine-tuning. The results on both tasks show that \model\ achieves substantial improvements over comparative foundation baselines, indicating its superior transferability and providing an advantageous initialization for subsequent fine-tuning stages. Subsequently, we fine-tune models for each dataset with available data. As shown in Table \ref{tab.fine-tune}, \model\ also consistently surpasses other competitors, demonstrating its robust generalization capabilities. 

\begin{figure*}[t]
    \centering
    \includegraphics[width=0.46\textwidth]{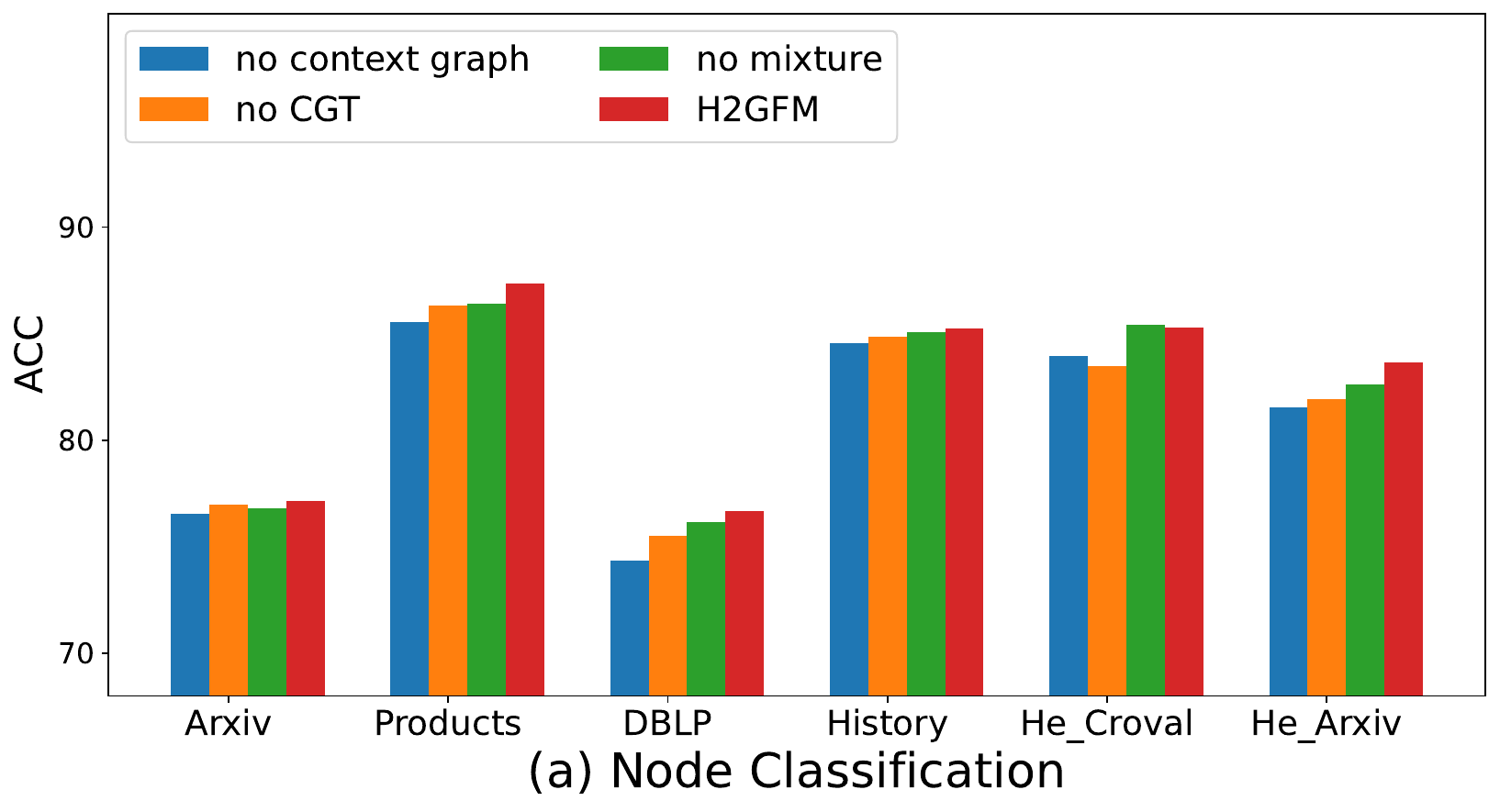}
    \hfill
    \includegraphics[width=0.46\textwidth]{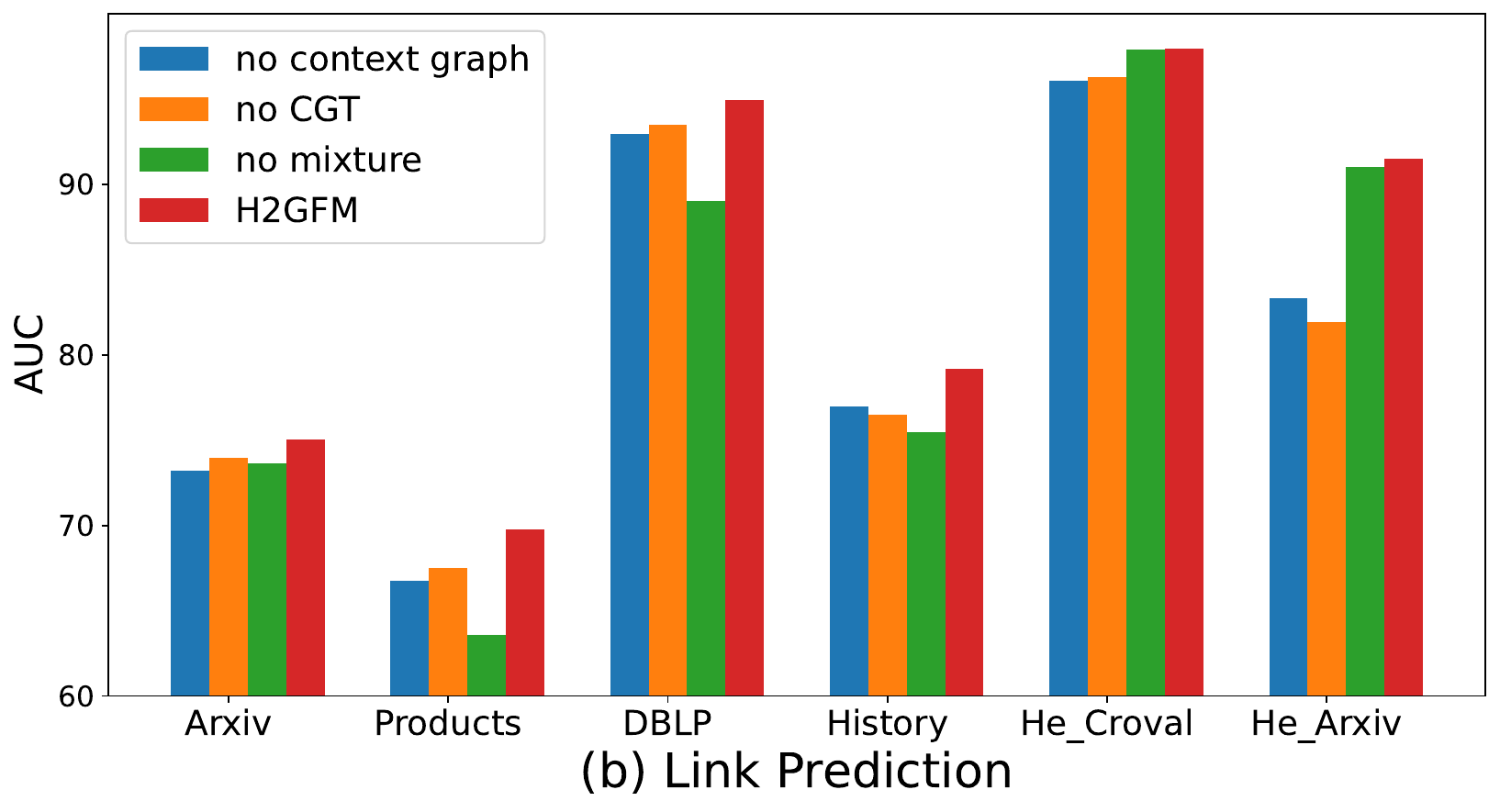}
    \caption{Ablation studies on co-training datasets.}
    \label{fig.ablation}
\end{figure*}

\subsection{Model Analysis}

\paragraph{Ablation study} We design subsequent ablation studies to investigate the effectiveness of each module in the proposed \model. (1) \textit{No context graph}: we sample  1-hop neighbors as in typical GNN aggregation, removing the usage of context graph and encoding. (2) \textit{No Context-Adaptive GT}: We replace our CGT by a conventional GT encoder, without taking high-order relation embeddings into consideration. (3) \textit{No mixture of CGT}: We remove the usage of a mixture of CGT experts and use only one CGT for all datasets. From the results reported in Fig.~\ref{fig.ablation}, we make the following observations. First, performance of \model\ drops when not using context graph, potentially due to the lack of diverse and high-order context information. Second, performance declines when CGT is not leveraged, suggesting its effectiveness in learning robust embeddings. It also worsens in case of no mixture of CGT, notably on link prediction, showing the importance of incorporating these modules to achieve optimal results.


\paragraph{Co-training effectiveness} To investigate whether it is beneficial to jointly train HoTAGs and HeTAGs, we conduct experiments on datasets from the same domain (CS paper): Arxiv (HoTAGs) and He\_Arxiv (HeTAG). We compare models' performance of training them individually versus co-training. As shown in Table. \ref{tab.co-ind}, OFA experiences performance drop during co-training, while \model\ demonstrates improvements on both datasets. This indicates the potential of co-training homogeneous and heterogeneous graphs from related domains, allowing them to complement each other. Furthermore, it is shown that \model\ can effectively leverage heterogeneous and domain-relevant information across graph data to boost performance.

\begin{wraptable}{r}{0.5\textwidth}
    \centering
    \small
    \caption{Evaluation of co-training effectiveness. ST indicates single training, while CT indicates co-training.}
    \label{tab.co-ind}
    \begin{tabular}{cccc}
        \toprule
        Methods & Setting & Arxiv & He\_Arxiv  \\
        \midrule
        \multirow{2}{*}{OFA} 
        & ST & 70.34 $\pm$ 0.33 & 78.41 $\pm$ 0.76 \\
        & CT & 70.03 $\pm$ 0.31 & 77.05 $\pm$ 0.68 \\
        \midrule
        \multirow{2}{*}{\model} 
        & ST & 76.92 $\pm$ 0.27 & 82.68 $\pm$ 0.76 \\
        & CT & \textbf{78.03} $\pm$ 0.78 & \textbf{85.61} $\pm$ 0.79 \\
        \bottomrule
    \end{tabular}
\end{wraptable}

 \vspace{-1mm}
\paragraph{Additional studies} We present additional  parameters analyses and scalability in Appendix F.


\section{Conclusion }

In this paper, we investigate the problem of developing GFM model to handle both homogeneous and heterogeneous TAGs, resulting in a novel model \model . Based on text meta-relation, we employ a simple yet effective context encoding to capture spatial and high-order semantic relationships in an unified latent space. We then introduce a mixture of context-adaptive graph transformers (CGT) to capture heterogeneous context and local structural patterns among graph types.
Finally, we conduct extensive experiments to demonstrate our model effectiveness under various scenarios. Future work can explore incorporating textual knowledge with geometrical information and symmetries from 3D graphs, addressing a wider range of applications in scientific domains such as chemistry, biology, and materials science. 

\textbf{Limitations.} In this work, we did not include graph-level tasks, such as graph classification, due to the lack of available benchmark datasets for heterogeneous graphs or HeTAGs. Nevertheless, these tasks can be readily adapted by applying a pooling layer for the output embeddings before passing to designated predictor.

\newpage
\bibliography{references}
\bibliographystyle{plain}


\appendix
\section*{Appendix}
\section{Algorithm and Complexity}

We outline the model training for \model\ in Algorithm~\ref{alg.c_train}.

\begin{algorithm}
\small
\caption{\textsc{Model Training for \model}}
\label{alg.c_train}
\begin{algorithmic}[1]
    \Require 
    Set of jobs $J^{tr}_i = \{D_j, T_k\}$, where $D_j \in$ graph datasets $D$, $D_j \in$ task sets $T$
    
    \Ensure \model\ model parameters $\theta$.
   \State Initialize parameters  $\theta$

    \While{not converged}
     \State sample mini batches for all jobs;
        \For{each mini batch $b$}
             \For{each target node $u$ in $b$}
                \For{each layer $l$}
                    \State Sample $N$ context neighbors $v$ from meta-paths $p$ 
                    \State Obtain unified $\vec{h}^l_{v}$ and $\vec{h}^l_{p}$ by Eqs (1), (3); 
                    \State Select $TopK$ experts by Eqs. (10), (11), (12); 
                    \For{each expert $i$ in $TopK$}
                        \State Calculate $CGT_i(\vec{h}^l_{u})$ by Eqs (4)-(8);
                    \EndFor
                \EndFor
                \State Calculate final $\vec{h}_u$ by Eq.(12);
            \EndFor
            \State Calculate the loss $\bL$ by Eq.(13);
            \State update $\Theta$ by minimizing $\bL$;
        \EndFor
    \EndWhile
    \State \Return $\theta$.
\end{algorithmic}
\end{algorithm}

In line 1, we initialize the model parameters. In line 3, we sample mini-batches of training data from all available jobs. From lines 4–12, robust node embeddings in each batch are calculated. Specifically, for each node $v$ in layer $l$, its context neighbors and meta-paths are sampled and their unified embeddings are retriveed in line 6-7. Next, the gating module selects top $k$ CGT experts to calculate node embeddings in line 8. Each expert, specialized in distinct structural patterns, calculates node embeddings in lines 9-11, then they are integrated in line 12. Finally, we compute the loss and update model parameters in line 14-15.

We first compare the complexity of one layer and one target node of a CGT versus standard GAT. For standard GAT, the aggregation for one node in the $l$-th layer has complexity $O(Nd^{l}d^{l-1}+Nd^{l})$, reducing to $O(Nd^{l}d^{l-1})$ for linear projection and attention computation, where $d^l$ is the output dimension of the $l$-th layer and $N$ is fixed neighbors number. For CGT, the attention needs one more linear projection for path embeddings and one more concatenation (Eq.4), thus complexity is still $O(Nd^{l}d^{l-1})$. Then, the computation of scaling and shifting vectors takes $O(Nd^{l}d^{l-1})$ based on Eqs.(7), and the aggregation step would take $O((N+1)d^ld^{l-1})$ based on Eqs.(8). The overhead to sample $N$ meta-paths of length $L_\text{max}$ is $O(NL_\text{max})$, which is trivial to the aggregation cost since $L_\text{max}$ is small. Then, the gating mechanism selects top $k$ experts, which cost $O(Kd^{l})$ where K is total expert number. The model employs top $k$ experts to calculate node embeddings, resulting in complexity $O(kNd^ld^{l-1})$. Since number of experts $k$ is small (8 in our setting), the total complexity can reduce to $O(Nd^ld^{l-1})$. Thus, while our model takes more computation, it still shares the same complexity class with a standard GAT. In practice, our model uses shallow CGTs, such as 1 layer, since long-range information can be captured by variable-length meta-paths in a layer. This compensates for the computational overhead by using mixture of experts .


\begin{figure*}[t]
    \centering
    \includegraphics[width=0.7\linewidth]{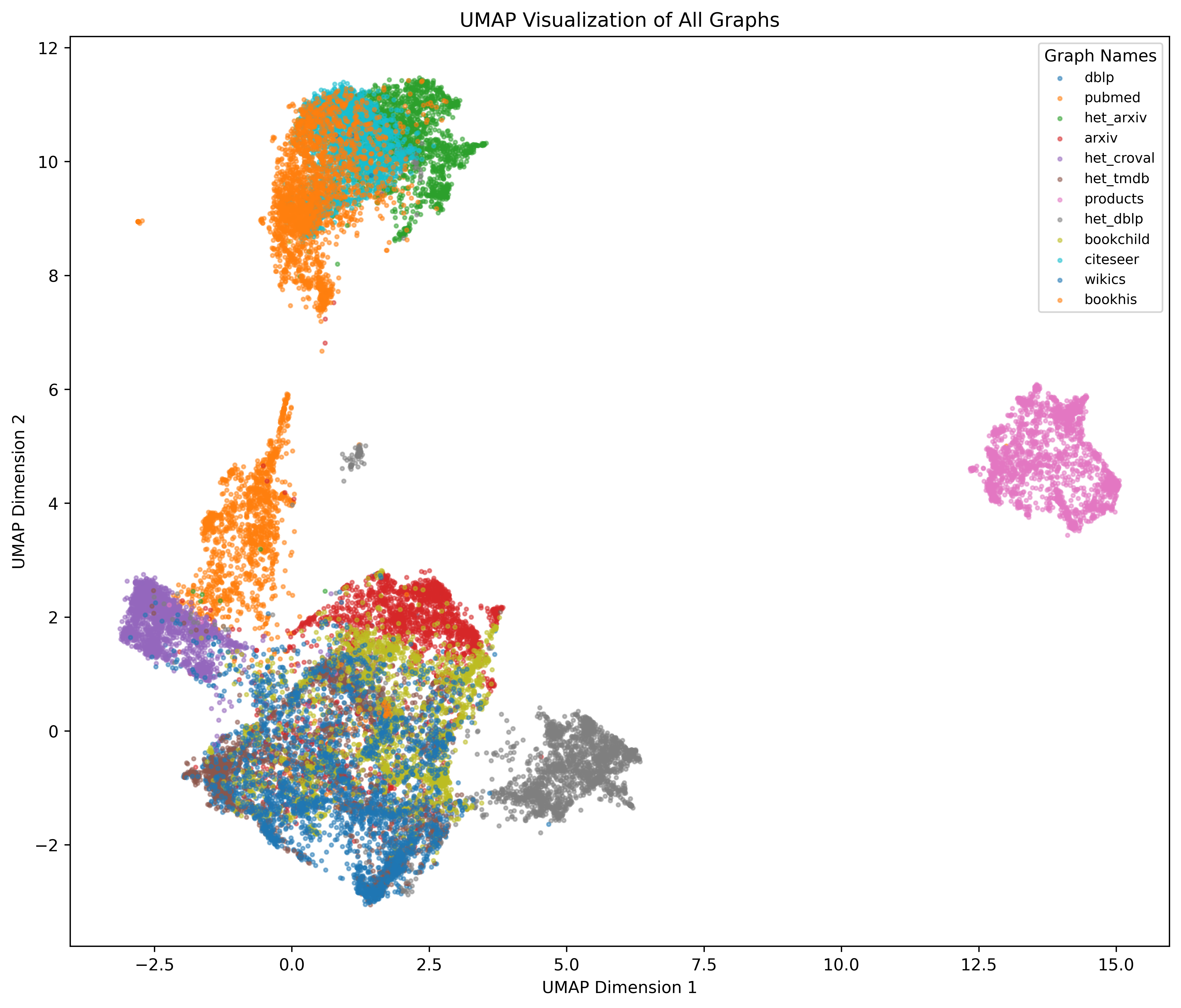}
    \caption{Visualizaion of node features of all datasets}
    \label{fig:viz}
\end{figure*}

\section{Details of Datasets}

In our experiments, we utilizes eight HoTAG and four HeTAG datasets from following sources \cite{chen2024text, liu2024multi}. Further data statistics and learning splits for node classification are provided in Table.\ref{tab:dat_node}. We also visualize the textual feature space of all datasets in Fig.\ref{fig:viz}

In HoTAG datasets, citation networks include Arxiv, Citeseer, DBLP, WikiCS (CS ), and Pubmed (BioMed),
where nodes are academic papers and edges represent citation relationships. E-commerce networks include Products, History (book of history topic), Child (book of children topic), where nodes are items and are linked if they are viewed or bought together. 

In HeTAG datasets, He\_Arxiv and He\_DBLP are academic networks consisting of node types "Author," "Paper," and "Field of Study". He\_CroVal is a question-answering network that includes "Question", "User" and "Tag" nodes. He\_TMDB is a movie collaboration network, comprising "Movie", "Director" and "Actor" nodes. We follow \cite{liu2024one} to utilize Gemini to generate text descriptions for their labels. The prompt template is:  "Generates short, formal descriptions for following topics (up to 50 words each topic), use the template: <topic>, <description>: <list of labels>".  

\begin{table*}
\centering
\small
\caption{Datasets details and splits for node classification.}
\label{tab:dat_node}
\begin{tabular}{lrrrrrrc}
\toprule
\textbf{Dataset} & \textbf{\# Nodes} & \textbf{\# Edges} & \textbf{Type} & \textbf{\# Train} & \textbf{\# Valid} & \textbf{\# Test} & \textbf{\# Classes} \\ 
\midrule
Arxiv & 169,343 & 2,315,598 & HoTAG & 135,473 & 16,935 & 16,935 & 40 \\
DBLP & 14,376 & 431,326 & HoTAG & 80 & 500 & 13,796 & 4 \\ 
Products & 316,513 & 19,337,722 & HoTAG & 25,321 & 6,330 & 284,862 & 39 \\
History & 41,551 & 503,180 & HoTAG & 24,930 & 8,310 & 8,331 & 12 \\
He\_CroVal & 44,386 & 164,981 & HeTAG & 980 & 1,242 & 31,931 & 6 \\ 
He\_ArXiv & 231,111 & 2,075,692 & HeTAG & 47,084 & 18,170 & 16,380 & 40 \\ 
\midrule
CiteSeer & 3,186 & 8,450 & HoTAG & 120 & 500 & 2566 & 6 \\
Pubmed & 19,717 & 88,648 & HoTAG & 60 & 500 & 19,157 & 3 \\
WikiCS & 11,701 & 431,726 & HoTAG & 580 & 1,769 & 5,847 & 10 \\
Child & 76,875 & 2,325,044 & HoTAG & 46,125 & 15,375 & 15,375 & 24 \\
He\_TMDB & 24,412 & 104,858 & HeTAG & 5,698 & 711 & 1,096 & 4 \\ 
He\_DBLP & 1,989,010 & 28,390,033 & HeTAG & 508,464 & 158,891 & 296,995 & 9 \\ 
\bottomrule
\end{tabular}
\end{table*}

\section{Details of Baselines}
\label{app:baselines}

In this section, we describe each baseline in more details.

\textit{GraphMAE} \cite{hou2022graphmae} is a self-supervised framework for learning node representations by masking and reconstructing node features. Using a GNN encoder, it processes visible features and graph structure to predict the masked parts, enabling effective pretraining without labels. This approach improves performance on downstream tasks, especially in limited data settings.

\textit{OFA} \cite{ofa} is a pioneer work on handling diverse graph classification tasks under a unified training paradigm. It first introduces Text-Attributed Graphs (TAG) to enable mapping node/edge textual features  into a shared language embedding space. Then they propose Nodes-of-Interest (NOI) mechanism and NOI subgraph to standardize node-level, link-level, and graph-level tasks into a single prediction format. To facilitate in-context learning, a Graph Prompting Paradigm (GPP) is employed by attaching task-specific prompt graphs to the input graph, including prompt nodes and class nodes described in natural language. These allow the model to adjust its predictions based on the task context and generalize across classification tasks  without retraining.

\textit{LLaGA} \cite{chen2024llaga} a Large Language and Graph Assistant that integrates LLM capabilities to handle graph-structured data, while maintaining the general-purpose nature of LLMs. Unlike approaches that convert graphs to verbose text descriptions, LLaGA reorganizes graph nodes into structure-aware sequences and maps these into the token embedding space through a versatile projector. The model excels in versatility, generalizability, and interpretability, allowing it to perform consistently across different datasets and tasks.

\textit{GFT} \cite{wang2025gft} is a cross-task, cross-domain graph foundation model that treats computation trees as transferable patterns in graphs. Unlike subgraph-based approaches, GFT leverages tree structures derived from message-passing processes and constructs a discrete tree vocabulary through vector quantization. The model operates in two phases: pre-training with tree reconstruction tasks to acquire general knowledge, and fine-tuning that unifies various graph tasks as tree classification. 

\textit{AnyGraph} \cite{xia2024anygraph} is a graph foundation model built on a Mixture-of-Experts (MoE) architecture to address cross-domain and in-domain heterogeneity in graph-structured data. To unify graphs with different structures and feature spaces, they first map varying adjacency matrices and node features into a fixed-size embedding space using SVD-based feature decomposition and further inject high-order connectivity by simplified GCN. Then a set of lightweight experts, typically MLP encoders, are employed for efficiency. The dynamic routing mechanism leverages contrastive learning signals to assign each input graph to the most appropriate expert.
It is regularized and periodically reprocessed to prevent expert dominance during training. 


\section{Model Settings}

For GraphMAE, we use hyperparameter settings in \cite{chen2024text}:
\begin{itemize}[leftmargin=0.5cm]
\item \textbf{Architecture:}  3-layer GAT, 768-dimensional embeddings with residual connection and batch norm normalization. Attention head is 4, feature and attention dropout are 0.5.  
\item \textbf{Pre-training:} Adam optimizer with learning rate 0.01 and weight decay 1e-5; the pre-train epochs is 20, and linear classifier is fine-tuned for each downstream task.
\end{itemize}

For reproducing OFA, we follow hyperparameter settings in \cite{chen2024text}:
\begin{itemize}[leftmargin=0.5cm]
\item \textbf{Architecture:}  5-layer RGCN, 768-dimensional embeddings with relu activation, neighbors sample per hop is 50. 
\item \textbf{Training:} Adam optimizer with learning rate 1e-3 an dropout ratio as 0.15; training epochs is 20 and batch size is 512.
\end{itemize}

For LlaGA, we employ a smaller base language model (Llama-3.2-1B) due to our limited resources, while maintaining its core architectural components. The following are the specific hyperparameters used in our reproduction:
\begin{itemize}[leftmargin=0.5cm]
    \item \textbf{Architecture:} Llama-3.2-1B as the backbone LLM (instead of Vicuna-7B in the original paper), with a linear projector to map graph embeddings to token space
    \item \textbf{Graph Encoding:} Hop-Field Overview Template (HO) with 4-hop neighborhood sampling, 10 neighbors per hop, SBERT for text feature encoding
    \item \textbf{Training:} AdamW optimizer with learning rate 2e-3, cosine learning rate scheduler with 3\% warmup ratio, weight decay 0, gradient checkpointing enabled
    \item \textbf{Implementation Details:} Maximum sequence length 4096, batch size 4, 1 training epoch, 
    \item \textbf{Task Configuration:} Model supports node classification and link prediction tasks across multiple datasets, with zero-shot transfer capabilities to new datasets
\end{itemize}

For AnyGraph, we follow the default setting for hyperparameters \cite{xia2024anygraph}:
\begin{itemize}[leftmargin=0.5cm]
\item \textbf{Architecture:} 3-layer GNN, 384-dimensional embeddings with layer norm normalization and LeakyReLU ratio as 0.5. The number of experts is 8.
\item \textbf{Training:} Adam optimizer with learning rate 0.0001 and weight decay 1e-7; training epochs is 50 and batch size 512. 
\end{itemize}

For GFT, we followed all the basic settings of GFT training, except that the output embedding dimension of 768 was changed to 384, which is suitable for our dataset. The following are the specific hyperparameters we used in the reproduction process.
\begin{itemize}[leftmargin=0.5cm]
\item \textbf{Architecture:} 2-layer GNN, 768-dimensional embeddings, ReLU with batch normalization.
\item \textbf{Tree Vocabulary:} 128 tokens of 768 dimensions each
\item \textbf{Pre-training:} Loss weights $\beta_1 = 10$, $\beta_2 = 100$, $\beta_3 = 1$, $\beta_4 = 0.01$; orthogonal regularizer $\lambda = 1$; AdamW optimizer with learning rate 1e-4 and weight decay 1e-5; 25 epochs; batch size 1024; edge and node feature drop rates of 0.2
\item \textbf{Fine-tuning:} Learning rates 5e-4 to 1e-3; maximum of 100 epochs; temperature scaling $\tau=1$; classifier weights 0.1 to 1 (varies by dataset)
\end{itemize}

\section{Environment}
All experiments are conducted on a workstation with a 24-core CPU, 64GB RAM, and an RTX A6000 GPU. We implemented \model\ using Python 3.10 and PyTorch 2.1 on Ubuntu-22.04.

\section{Addiontional results}

\subsection{Parameter studies}

\begin{wrapfigure}{R}{6cm}
\centering
\vspace{-4mm}
\includegraphics[width=0.95\linewidth]{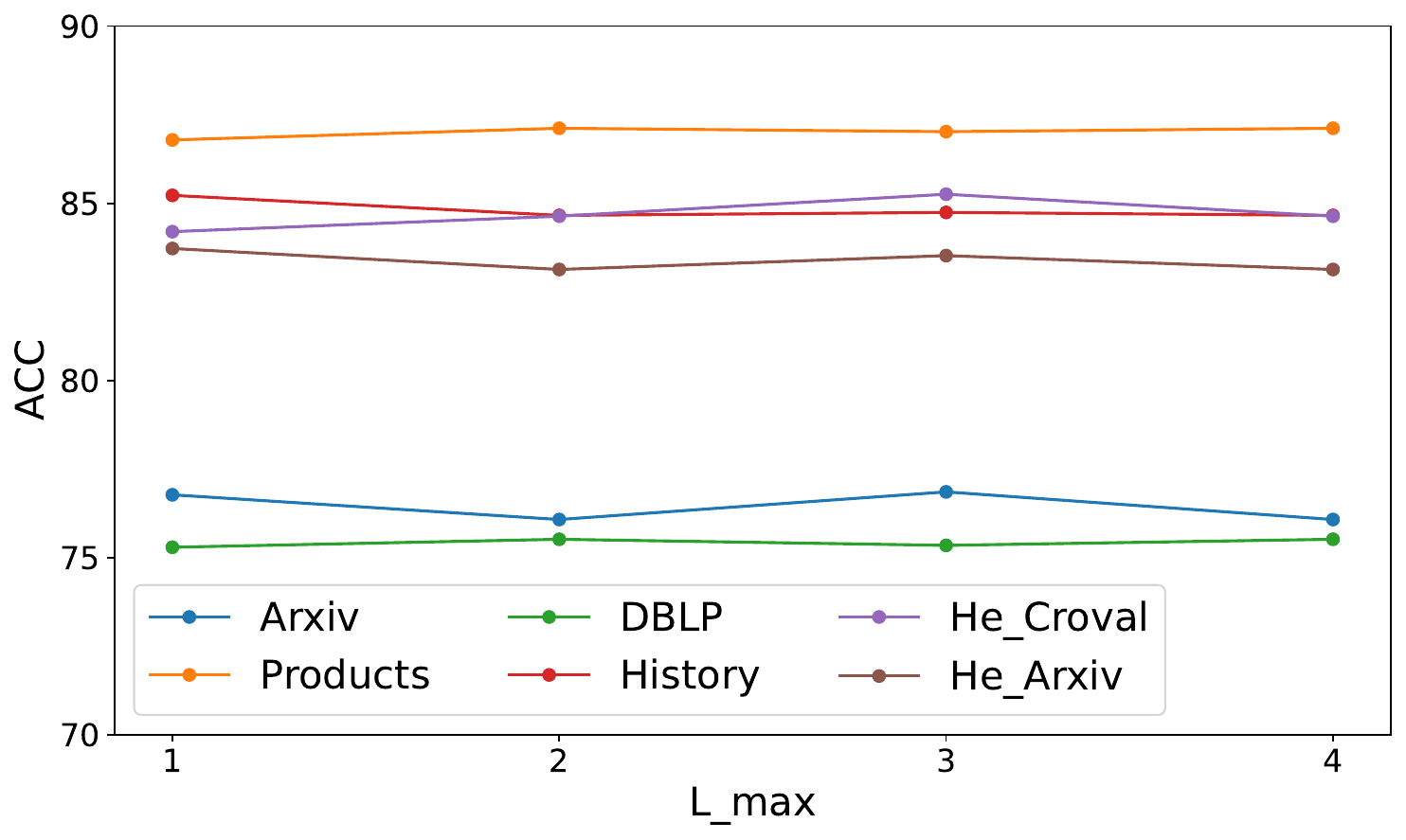}
\caption{Impact of maximum path length.}
\label{fig.lmax}
\end{wrapfigure}
We study the impact of maximum length $L_{max}$ for the random-walk meta-path sampling strategy. From Fig.\ref{fig.lmax}, while performance varies across datasets, larger values of $L_{max}$ tend to achieve more consistent results on a wide range of data, suggesting that incorporating long-range context can be beneficial for prediction. Although increasing $L_{max}$ would incur additional computation for path sampling and encoding, this overhead can be negligible compared to node embeddings computation and aggregation cost (Appendix. A).

\begin{figure*}[t]
    \centering
    \includegraphics[width=0.46\textwidth]{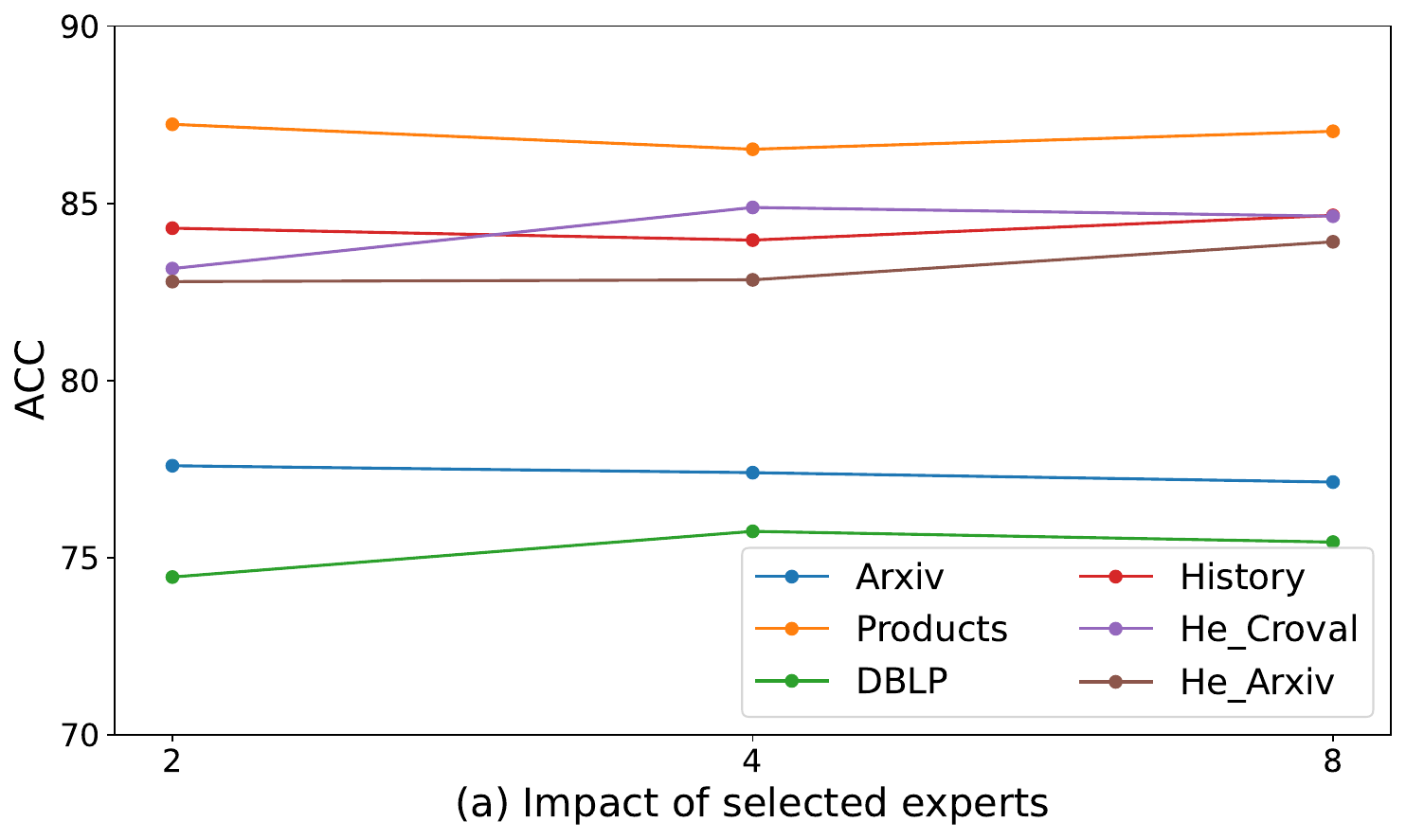}
    \hfill
    \includegraphics[width=0.46\textwidth]{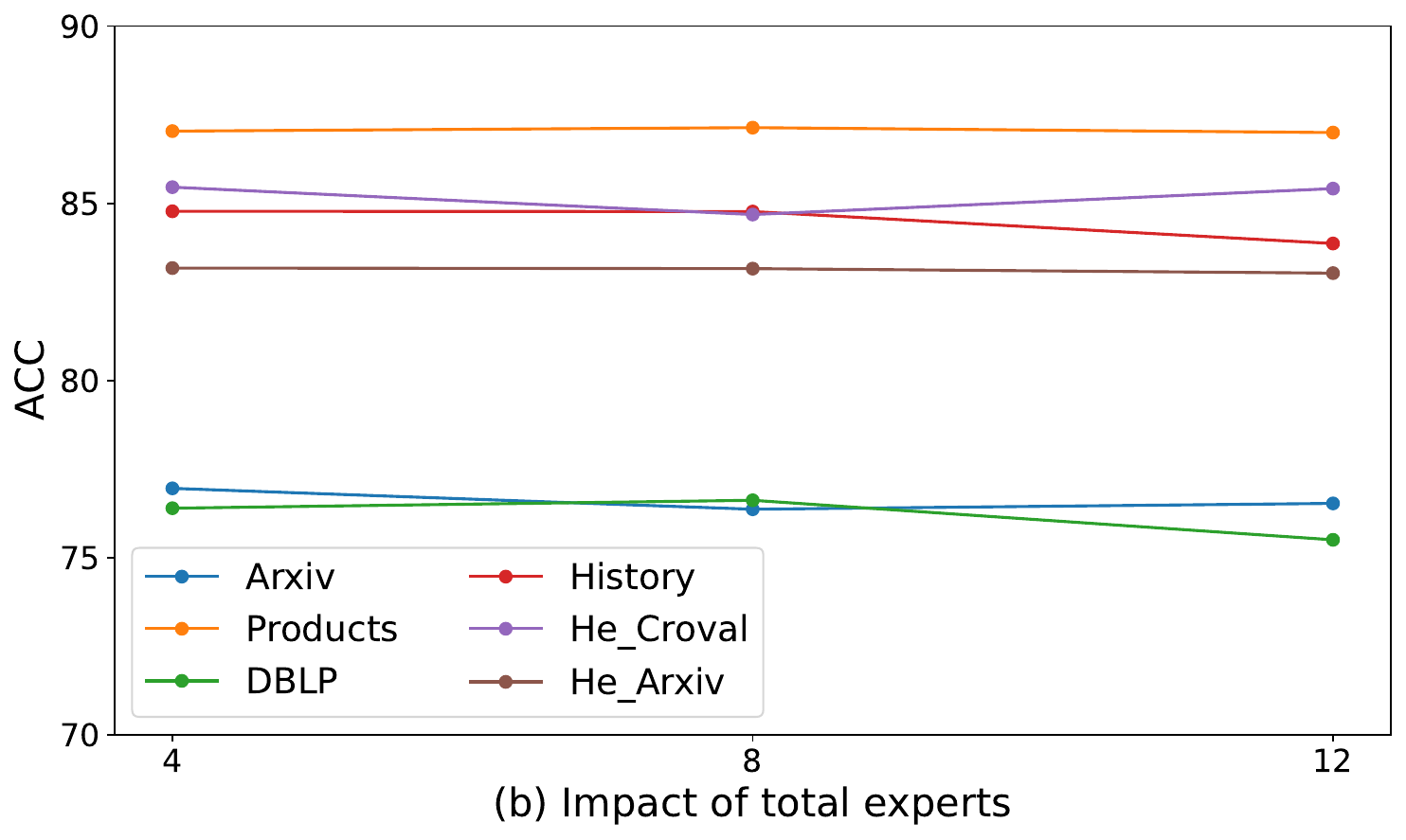}
    \caption{Parameter studies.}
    \label{fig.param}
\end{figure*}
We also investigate the impact of two key hyperparameters for mixture of CGT experts, including the number of selected experts $k$ and the total number of experts $n$. As shown in Fig.\ref{fig.param} (a), increasing activated experts generally boosts the performance but  also raises computation cost for inference. Thus, selecting an appropriate value $k$ should be considered to balance the trade-off between effectiveness and efficiency. We then fix $k$=4 to examine the impact of the total number of experts $n$. From Fig.\ref{fig.param} (b), results remain consistent across different values of $n$. Increasing total numbers may compromise the performance in some cases, potentially due to the challenges in maintaining load balancing.  

\subsection{Scalability studies}

To assess model scalability, we sample five subgraphs from the dataset Arxiv, where sizes evenly range from 20k to 100k nodes. We present the training time for one epoch of \model\ on these subgraphs in Table~\ref{table.time}. We observe the training time grows linearly with graph size in terms of the number of nodes, demonstrating \model\ potentials for scaling to large datasets. We also report the model size in Table.\ref{table.size}. The number of \model\ trainable parameters is comparable to other graph-based baselines, indicating its parameter efficiency. Among that, LlaGA requires a significantly larger number of parameters due to its reliance on a large language model (LLM) for prediction.

\begin{figure}[tbp]
\begin{minipage}{0.5\linewidth}
\addtolength{\tabcolsep}{2pt}
\captionof{table}{Training time for one epoch.\label{table.time}}
\centering
\begin{tabular}{rrr}
\toprule
\textbf{Nodes} & \textbf{Edges} & \textbf{Time} \\
\midrule
20k & 32k & 1.82s \\
40k & 128k & 3.67s \\
60k & 296k & 5.42s \\
80k & 540k & 7.39s   \\
100k & 836k & 9.26s \\
\bottomrule
\end{tabular}
\end{minipage}%
\begin{minipage}{0.45\linewidth}
\centering
\captionof{table}{Model learnable parameters.\label{table.size}}
\begin{tabular}{l|r}
\toprule
\textbf{Methods} & \textbf{\# Params} \\
\midrule
OFA & 21.5M  \\
LlaGA & 1B+ \\
AnyGraph & 9.5M \\
GFT & 7M \\
\model & 16M \\

\bottomrule
\end{tabular}

\end{minipage}
\end{figure}

\section{Data and Code}
We provide the code of \model\ and instruction for downloading data in supplementary materials.

\end{document}